\pdfoutput=1
\documentclass{article}

\PassOptionsToPackage{numbers, compress}{natbib}

\usepackage[preprint]{neurips_2023}

\usepackage[utf8]{inputenc} 
\usepackage[T1]{fontenc}    
\usepackage{hyperref}       
\usepackage{url}            
\usepackage{booktabs}       
\usepackage{amsfonts}       
\usepackage{nicefrac}       
\usepackage{microtype}      
\usepackage[table,xcdraw]{xcolor}         

\usepackage{pifont}
\usepackage{lipsum}
\usepackage{subcaption}
\usepackage{wrapfig}
\usepackage{multirow}
\usepackage{makecell}
\usepackage{xspace}
\usepackage{enumitem}
\usepackage{amsmath}
\usepackage{listings}
\usepackage{algorithm}
\usepackage{algorithmic}
\usepackage{amsfonts,amssymb}
\usepackage{mathrsfs}
\usepackage{subcaption}
\usepackage{natbib}
\setcitestyle{numbers,square}
\usepackage{graphicx}       
\usepackage{cleveref}
\usepackage{listings}
\usepackage[most]{tcolorbox}

\newcommand{\hhide}[1]{}
\newcommand{\hide}[1]{}
\newcommand{\model}[0]{GLM-4-Voice\xspace}

\title{GLM-4-Voice: Towards Intelligent and Human-Like End-to-End Spoken Chatbot}

 \author{
 \centerline{
\bf Aohan Zeng$^{\ddagger\S*}$, Zhengxiao Du$^{\ddagger\S*}$, Mingdao Liu$^{\ddagger}$, Kedong Wang$^\S$, Shengmin Jiang$^{\S}$, Lei Zhao$^\S$
}\\
\centerline{
\bf Yuxiao Dong$^\ddagger$, Jie Tang$^{\ddagger}$
}
\\
\centerline{$^{\S}$Zhipu.AI\qquad$^\ddagger$Tsinghua University}\\
\url{https://github.com/THUDM/GLM-4-Voice}\\
{\includegraphics[height=3.5ex]{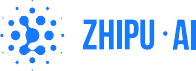}}
}

\begin{document}

\maketitle
\renewcommand{\thefootnote}{\fnsymbol{footnote}}
    \footnotetext[1]{Equal contribution. Email: \texttt{\{zah22,zx-du20\}@mails.tsinghua.edu.cn}}
    \footnotetext[4]{Work was done when ML, LZ interned at Zhipu.AI.}
\renewcommand{\thefootnote}{\arabic{footnote}}

\begin{abstract}
We introduce \model, an intelligent and human-like end-to-end spoken chatbot. It supports both Chinese and English, engages in real-time voice conversations, and varies vocal nuances such as emotion, intonation, speech rate, and dialect according to user instructions.
\model uses an ultra-low bitrate (175bps), single-codebook speech tokenizer with 12.5Hz frame rate derived from an automatic speech recognition (ASR) model by incorporating a vector-quantized bottleneck into the encoder.
To efficiently transfer knowledge from text to speech modalities, we synthesize speech-text interleaved data from existing text pre-training corpora using a text-to-token model. We continue pre-training from the pre-trained text language model GLM-4-9B with a combination of unsupervised speech data, interleaved speech-text data, and supervised speech-text data, scaling up to 1 trillion tokens, achieving state-of-the-art performance in both speech language modeling and spoken question answering. We then fine-tune the pre-trained model with high-quality conversational speech data, achieving superior performance compared to existing baselines in both conversational ability and speech quality. The open models can be accessed through \url{https://github.com/THUDM/GLM-4-Voice} and \url{https://huggingface.co/THUDM/glm-4-voice-9b}.
\end{abstract}

\section{Introduction}

The success of large language models (LLMs) has driven significant advancements in conversational AI, enabling the development of text-based chatbots and digital assistants. However, LLMs are primarily designed to process text input and generate text output, focusing on semantic and logical communication. In contrast, human communication extends beyond semantics, often conveying emotions and subtle nuances. Voice-based interaction, therefore, provides a more natural and intuitive medium for human-computer interaction, offering richer and more engaging user experiences. Traditional spoken chatbot typically rely on a pipeline combining Automatic Speech Recognition (ASR), LLM processing, and Text-to-Speech (TTS) synthesis. While functional, this approach is often hindered by high latency, compounded errors introduced during the ASR and TTS stages, and a limited capacity to capture and express emotional nuances.

Speech-language models (SpeechLMs), which process both speech input and output in an end-to-end manner, offer a promising approach for building spoken chatbots. Efforts such as \cite{GSLM, TWIST} have explored pre-training on speech data in a manner similar to large language models (LLMs). Similarly, \citet{moshi} scaled speech data to 7 million hours for model training. However, these approaches face a significant limitation: the relative scarcity of speech data compared to the extensive text corpora available online. This data imbalance makes it challenging to fully leverage the capabilities of text-based LLMs, ultimately constraining the intelligence of SpeechLMs. Other methods aim to align speech and text modalities~\cite{llama-omni, mini-omni} by integrating a speech encoder and a text-to-speech module into existing LLMs and fine-tuning them on spoken dialogue datasets. While this approach provides a straightforward way to develop speech-to-speech models from LLMs, it lacks the ability to deliver truly human-like speech output due to the absence of dedicated speech pre-training. This limitation hinders these models from capturing the rich nuances and expressiveness inherent in human speech.

In this paper, we introduce \model, an intelligent and human-like spoken chatbot. We use a single code-book supervised speech tokenizer with 12.5Hz frame rate to efficiently represent speech. A flow-matching-based speech decoder is employed to convert speech tokens into natural-sounding speech. To bridge the gap between text and speech modalities, we conduct large-scale speech-text pre-training using 1 trillion tokens. This includes synthetic interleaved speech-text corpora derived from text pre-training data, as well as unsupervised speech data and supervised speech-text datasets (e.g., ASR and TTS). The resulting base model demonstrates strong performance across various tasks, including speech language modeling, spoken question answering, ASR, and TTS. To further enhance the chatbot’s conversational capabilities, we fine-tune the base model on high-quality conversational datasets using a "streaming thoughts" template. This template alternates between outputting text and speech tokens, improving the model's ability to generate seamless, low-latency responses while maintaining high-quality performance.

\section{Related Work}
\subsection{Speech Tokenization}
Speech tokenizers, which transform a audio clip into discrete tokens, can be categorized into two directions. The neural acoustic codecs~\citep{SoundStream,neuralac,audio_rvqgan,wavtokenizer} target at reconstructing high-quality audio at low bitrates. The semantic tokens~\citep{hubert,w2vbert} are extracted from speech representations learned with self-supervised learning on speech data. Recently, SpeechTokenizer~\citep{speechtokenizer} and Mini~\citep{moshi} unify semantic and acoustic tokens as different residual vector quantization (RVQ) layers, but they also suffer from multiple tokens at the same position, leading to either parallel prediction of semantic and acoustic tokens, or degradation to semantic tokenizers for language models. CosyVoice~\citep{cosyvoice} proposes the supervised semantic tokenizer derived from a speech recognition model, and successfully apply the tokenizer to text-to-speech synthesis. The application of the tokenizer on speech language modeling is not explored.

\subsection{Speech Language Modeling}
Speech language models are autoregressive models pretrained on unsupervised speech data.
\citet{GSLM} first proposes generative spoken language modeling (GSLM), which trains the next-token-prediction objective on discrete semantic tokens produced by self-supervised learning. AudioLM~\citep{AudioLM} proposes a hybrid tokenization scheme that combines these semantic tokens with acoustic tokens from a neural audio codec~\citep{SoundStream}. TWIST~\citep{TWIST} trains the speech language model using a warm-start from the pretrained text language model OPT~\citep{OPT}. Moshi~\cite{moshi} scales up the size of natural speech data in TWIST to 7 million hours. Spirit-LM~\citep{spiritlm} further extends TWIST by adding speech-text interleaving data curated from speech-text parallel corpus. However, the scarcity of speech-text parallel corpus restricts the scale of interleaving data. 

\subsection{End-to-End Spoken Chatbots}
Early works in speech-to-speech models mainly focus on processing tasks like speech translation~\citep{speechnet,speecht5}. Since success of ChatGPT in text-based chatbots, many works have explored methods to develop speech-based chatbots that can understand and respond in speech. SpeechGPT~\citep{speechgpt} proposes to combine existing large language models (LLM) with discrete speech representations to obtain speech conversational abilities.
Moshi~\citep{moshi} proposes a full-duplex spoken dialogue framework based on their pretrained speech language model.
Qwen-Audio~\citep{qwen-audio} adapts pre-trained textual language models for speech understanding by aligning speech representations of the Whisper~\citep{whisper} encoder. The model can understand speech, but not generate speech. Llama-Omni~\citep{llama-omni} and Freeze-Omni~\citep{freezeomni} extend the method by adding a text-to-speech model after the language model to transform the text output to speech output. In this way language models can only control the content of speech, but not the styles and prosodies. Mini-Omni~\citep{mini-omni} directly fine-tunes language models to generate text and speech responses simultaneously with only instruction datasets. Without speech pre-training, the quality of both text and speech responses is severely limited, as we will show in the experiments.
\section{Architecture}
In this section, we introduce the architecture of \model. Our goal is to build a human-like, end-to-end spoken chatbot with high intelligence. To achieve this, the model must 1) comprehend the user’s speech and provide a semantically accurate response, and 2) follow the user’s spoken instructions, generating speech with paralinguistic features that meet the user's expectations. Inspired by the successful pre-training and fine-tuning paradigm used in LLMs, we believe that these capabilities for spoken chatbots can be best developed through extensive pre-training on diverse speech corpus, rather than simply fine-tuning existing LLMs with speech question-answering data, as in recent spoken chatbot approaches~\citep{llama-omni,mini-omni}.

To achieve this goal, \model is designed with minimal modifications to the auto-regressive transformer architecture. For speech tokenization, we utilize a supervised speech tokenizer, which effectively captures semantic information at a ultra-low bitrate (175bps) while maintaining high-quality speech reconstruction. Additionally, we adopt a single-codebook approach for speech tokenization, avoiding the complex architectural adjustments often required for multi-layer speech token generation~\citep{moshi,mini-omni}. This approach helps preserve the model’s text processing capabilities while enabling efficient speech modeling. Furthermore, the model employs a unified speech representation for both input and output, enabling next-token prediction for speech data and facilitating efficient pre-training on unsupervised speech corpora.

We use the same speech tokenizer and speech decoder as described in \citet{speechpretrain}. To enable low-latency interaction, we adapt the speech decoder to support streaming inference and design a \textit{streaming thought} template capable of alternating between text and speech tokens during the supervised fine-tuning stage, as detailed in~\Cref{sec:inference} and~\Cref{sec:speech-decoder}.

 
\subsection{Speech Tokenizaion}
\begin{wrapfigure}{r}{0.5\textwidth}
    \centering
    \vspace{-1em}
    \includegraphics[width=0.5\textwidth]{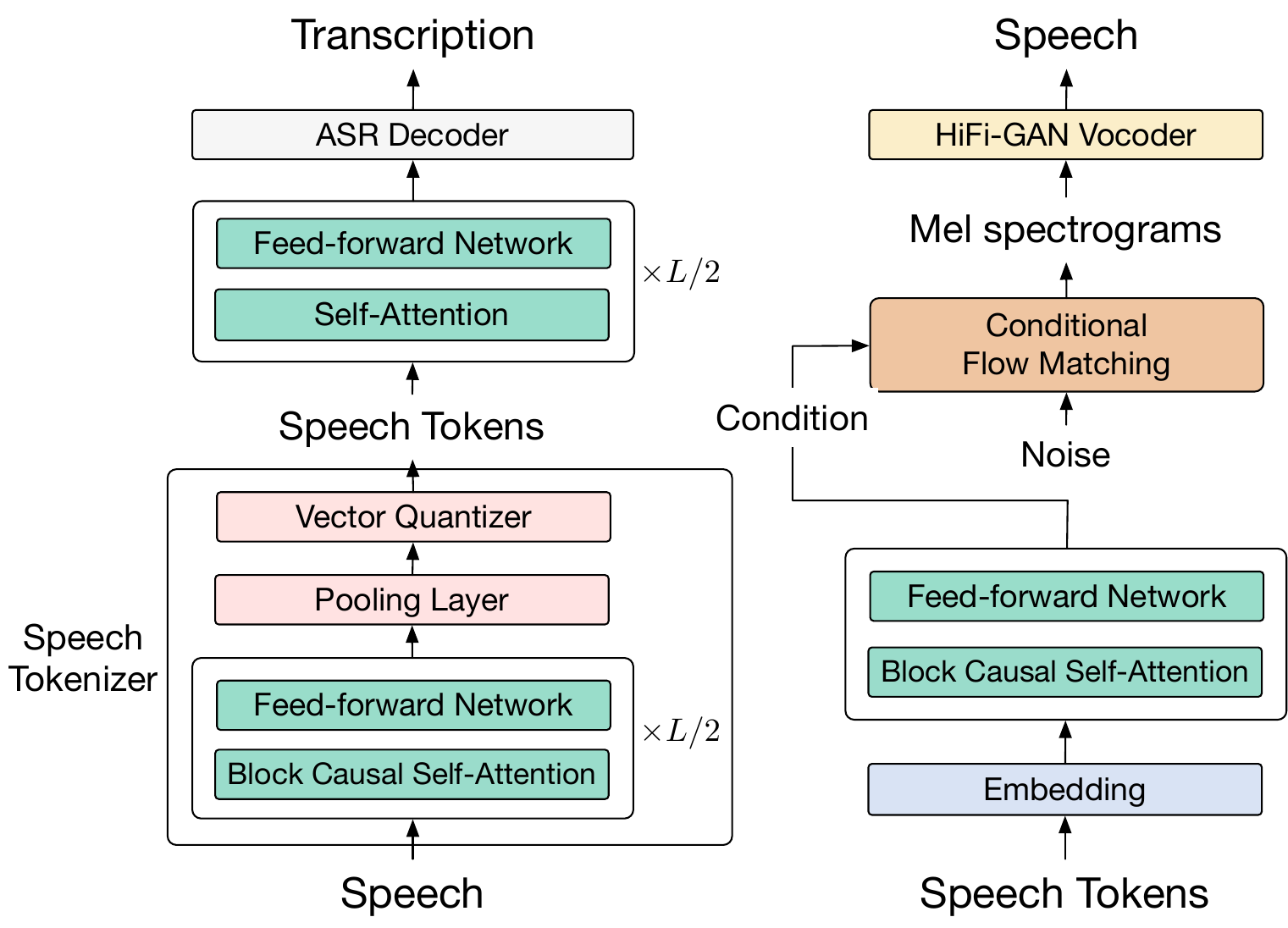}
    \label{fig:tokenizer-arch}
    \vspace{-1em}
    \caption{Architecture of the Speech Tokenizer and Speech Decoder for \model.}
    \vspace{-0.5em}
\end{wrapfigure}
\label{sec:speech-tokenization}
The speech tokenizer converts continuous waveforms into discrete speech tokens, which reserve semantic information and a part of acoustic information. Previous methods can be categorized into two directions. Acoustic tokenizers are trained with reconstruction/adversarial objectives of speech waveform. Acoustic tokens reserve enough information to reconstruct the original audio, but to represent the additional information it relies on either high sampling rate (i.e. number of tokens per second) or residual vector quantization~\cite{SoundStream} (i.e. multiple stacked codebooks).
Semantic tokens are extracted from self-supervised representations learned on automatically discovered speech units~\cite{hubert}. Semantic tokens discard additional information that is unnecessary to represent semantic meaning of speech, but also result in low-quality speech synthesis and a loss of acoustic details~\cite{Expresso}.  The ideal speech tokenizer for speech-text language modeling should have several key features: 1) low sampling rate with a single codebook to support autoregressive generation. 2) aligning with texts to transfer knowledge of pretrained language models. 3) support of high-quality speech synthesis. 

We adopt the 12.5Hz speech tokenizer variant described in \citet{speechpretrain}. To make the paper self-contained, we briefly describe the architecture of the speech tokenizer. Inspired by the supervised semantic tokenizer in text-to-speech synthesis~\citep{cosyvoice}, we finetune a pretrained automatic speech recognition model (we use \textit{whisper-large-v3} in the Whisper family~\citep{whisper}) with an additional pooling layer and a vector quantization layer~\citep{vqvae} in the middle of the encoder.
The codebook vectors are learned with exponential moving average (EMA) and we reset vectors whose mean usage falls below a certain threshold with randomly-selected continuous representations before quantization to overcome codebook collapse following \citet{jukebox}.

\paragraph{Causality for Streaming Inference}
To enable streaming encoding of input speech during inference, we adapt the architecture of Whisper encoder to introduce causality~\cite{speechpretrain}. Specifically, we replace the convolution layer before the encoder Transformer with causal convolution~\citep{wavenet}. We also replace the bidirectional attention in the encoder with block causal attention.

\begin{table}[t]
    \setlength{\tabcolsep}{3pt} 
    \centering
    \caption{\textbf{Evaluation results of speech tokenizers and decoders.} LS stands for LibriSpeech. Evaluation on LibriSpeech (English) is measured using word error rate (WER), while AISHELL-1 (Chinese) is evaluated using character error rate (CER). We fine-tuned the ASR model \texttt{whisper-large-v3} with vector quantization and various pooling layers to create tokenizers with different sampling rates. For further development of \model, we selected the 12.5 Hz variant.}
    \resizebox{\textwidth}{!}{
    \begin{tabular}{@{}lrrrrrrrr@{}}
    \toprule
     & \textbf{Frame} & \textbf{BitRate} & \multicolumn{3}{c}{\textbf{ASR$\downarrow$}} & \multicolumn{3}{c}{\textbf{Reconstruction}} \\
    & \textbf{Rate} & \textbf{(bps)} & LS-clean & LS-other & AISHELL-1 & WER$\downarrow$ & VisQOL$\uparrow$ & MOSNet$\uparrow$ \\
    \midrule
    SpeechTokenizer & 50Hz & 1.50K & $\emptyset$ & $\emptyset$ & $\emptyset$ & 9.97 & 1.53 & 2.67 \\
    SpeechTokenizer & 50Hz & 4.00K & $\emptyset$ & $\emptyset$ & $\emptyset$ & 6.32 & 3.07 & 3.10 \\
    Moshi (Mimi) & 12.5Hz & 1.10K & $\emptyset$ & $\emptyset$ & $\emptyset$ & 8.36 & 2.82 & 2.89 \\
    \midrule
    whisper-large-v3 & 50Hz & - & 2.50 & 4.53 & 9.31 & $\emptyset$ & $\emptyset$ & $\emptyset$ \\
    SenseVoice-Large & 50Hz & - & 2.57 & 4.28 & 2.09 & $\emptyset$ & $\emptyset$ & $\emptyset$ \\
    \midrule
    GLM-4-Voice-Tokenizer
    & 12.5Hz & 175 & 2.10 & 4.90 & 3.02 & 8.43 & 2.52 & 3.39 \\
    & 50Hz & 600 & 1.85 & 3.78 & 2.70 & 6.24  & 2.67 &3.38 \\
    & 25Hz & 300 & 1.94 & 4.16 & 2.86 & 6.80 & 2.60 & 3.33 \\
    & 6.25Hz & 100  & 14.41  & 2.34 & 3.24  & 14.41 & 2.34 & 3.24  \\
    \bottomrule
    \end{tabular}
    }
    \label{tab:tokenizer_asr}
\end{table}

\paragraph{Training Details} We fine-tune the vector-quantized Whisper model with a collection of ASR datasets, including LibriSpeech~\citep{librispeech}, GigaSpeech~\citep{GigaSpeech}, MLS-Eng~\citep{MLS}, Wenet~\citep{wenet}, CommonVoice~\citep{commonvoice}, AISHELL-1~\citep{aishell1}, and a proprietary Chinese ASR dataset of 10k hours. We also include 700k hours unsupervised speech data with pseudo labels generated by \textit{whisper-large-v3}~\cite{whisper} for English and \textit{paraformer-large}~\cite{funasr} for Chinese. All of our speech tokenizers are fine-tuned from \textit{whisper-large-v3} for 2 epochs with batch size 4096 and learning rate 1e-5. The ratio of supervised samples to pseudo-labeled samples is 1:3. The codebook vectors are updated with exponential moving average with decay coefficient 0.99 and the commitment loss coefficient is 10.0. To reduce the information loss of average pooling, we increase the codebook size as the sampling rate decreases.

\paragraph{Evaluation} We measure the reservation of semantic information in the speech tokens by the accuracy of the finetuned ASR model. The results on LibriSpeech~\citep{librispeech} and AISHELL-1~\citep{aishell1} are shown in \Cref{tab:tokenizer_asr}, with \textit{whisper-large-v3}~\cite{whisper} and \textit{SenseVoice-Large}~\cite{funasr} as baselines. Overall all the tokenizers reserve enough semantic information to achieve accurate ASR performance. Considering the reconstruction results in the following section, we select the 12.5Hz tokenizer for \model.

\subsection{Speech Decoder}
\label{sec:speech-decoder}
The speech decoder synthesizes speech waveforms from discrete speech tokens and is crucial for ensuring the quality and expressiveness of generated speech. To minimize latency during speech interaction, the decoder must also support streaming inference. As in \citet{speechpretrain}, we adopt the decoder architecture of CosyVoice~\cite{cosyvoice}, which comprises a speech token encoder, a conditional flow matching model~\cite{mehta2024matcha}, and a HiFi-GAN vocoder~\cite{hifi_gan}.

\paragraph{Training Details} We train the speech token encoder and the flow matching model from scratch, with a two-stage training paradigm to fully utilize the abundant speech data of varied quality. During the pre-training stage, we use all the speech samples in the unsupervised speech data of various speakers and quality. During the fine-tuning stage, we use high-quality speech samples from a single speaker. 

\paragraph{Support for Streaming Inference} To enable streaming inference and reduce latency, we incorporate truncated audio samples (i.e., the first $n \cdot b$ seconds of the audio, where $n = 1, 2, 3, \ldots$, and $b$ is the block size) during the fine-tuning stage. This prepares the model to handle streaming scenarios effectively. During inference, the decoder processes speech tokens corresponding to the first $n \cdot b$ seconds of audio. It uses the speech from the initial $(n-1)b$ seconds as the prompt and predicts the speech content from $(n-1)b$ to $n \cdot b$ seconds. This approach allows the model to generate speech tokens with a minimum delay of $b$ seconds. Based on empirical studies, we set $b = 0.8$ for \model, which implies that at least 10 speech tokens are required to generate the initial speech output.

\paragraph{Evaluation}
We take the reconstruction results from \citet{speechpretrain} to demonstrate the performance of our speech decoder with low-bit-rate speech tokens. We evaluate our speech decoder on speech reconstruction of LibriSpeech~\cite{librispeech}. 
and compare our tokenizer with SpeechTokenizer~\cite{speechtokenizer} and Mini~\cite{moshi}. Following \citet{moshi}, we also evaluate a variant of SpeechTokenizer that only keeps the first 3 RVQ layers to obtain a 1.5kbps bitrate. \Cref{tab:tokenizer_asr} shows that our speech decoder performs well across various sampling rates, with the 12.5Hz variant offering an optimal balance between efficiency and quality. It maintains high quality scores (MOSNet 3.39) and content preservation (WER 8.43) while significantly reducing bitrate (175).


\subsection{Inference}
\label{sec:inference}
\paragraph{Decoupling Speech-to-Speech Task}

\begin{figure}[t]
    \centering
    \includegraphics[width=\linewidth]{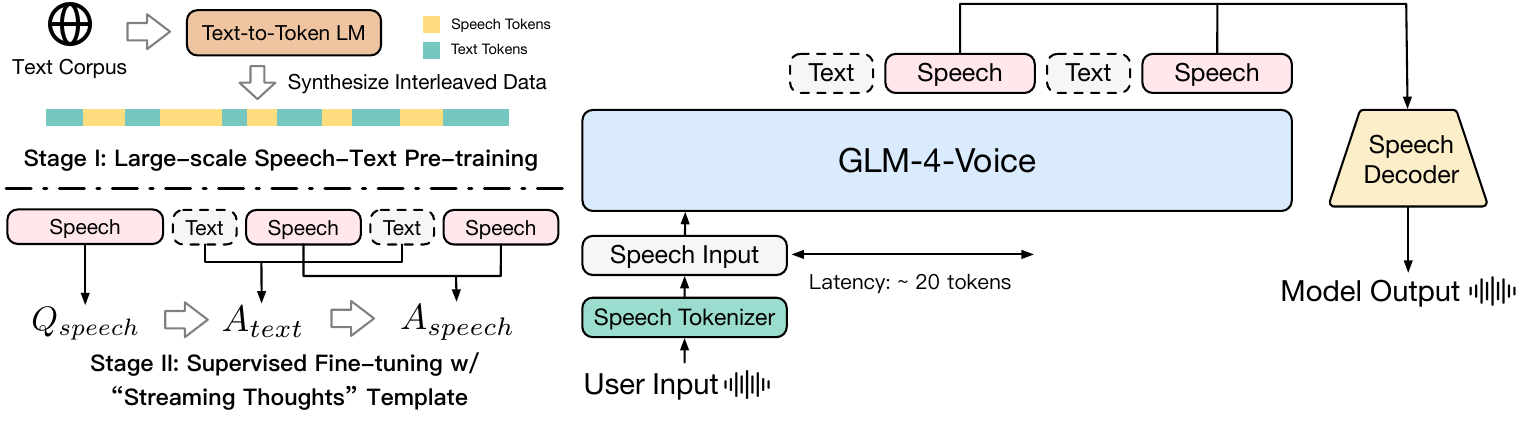}
    \caption{\textbf{Left:} Data construction of two training stage of \model.  \textbf{Right:} Model architecture of \model.}
    \label{fig:arch-overview}
\end{figure}

An ideal speech language model would operate solely on speech tokens for direct speech-to-speech tasks. However, given the success of large language models and the assumption that text representing the semantic content of most speech, we decouple the speech-to-speech task into two sub-tasks: speech-to-text and speech-and-text-to-speech. Given the user's speech input $Q_s$, the correspond text response $A_t$, and the speech output $A_s$, these tasks are defined as follows:

\begin{itemize}[leftmargin=*,itemsep=0pt,parsep=0.2em,topsep=0em,partopsep=0em]
\item \textbf{Speech-to-Text}: The model generates a text response, $A_t$, based on the user's speech input, $Q_s$.
\item \textbf{Speech-and-Text-to-Speech}: Leveraging both $Q_s$ and $A_t$, the model generates spoken output, $A_s$, with adaptive tone and prosody to ensure conversational coherence.
\end{itemize}

We adopt the decoupling strategy for the inference process. First, the model generates the text answer $A_t$ based on the user input $Q_s$, and then generates $A_s$ using both $Q_s$ and $A_t$.
In this way the generation of speech response $A_s$ is guided by the text response $A_t$ to improve performance.
However, this approach results in a high initial token delay, as it requires waiting for the complete generation of $A_t$ before starting on $A_s$. To address this, we apply a template named called \textit{Streaming Thoughts}. As illustrated in~\Cref{fig:arch-overview}, given $Q_s$, the model alternates between outputting text and speech tokens at a specified ratio, which are then concatenated to form $A_t$ and $A_s$, respectively. Specifically, based on our 12.5Hz tokenizer, we alternate between generating 13 text tokens and 26 speech tokens. This 1:2 ratio is chosen to ensure that text generation is consistently faster than speech. Otherwise, the generated speech tokens would lack the necessary context from the text tokens. The choice of 26 speech tokens is based on empirical observations, allowing the model to produce a coherent portion of content before synthesizing it to ensure accuracy in the synthesized speech.

\paragraph{Overall Latency} 

The overall response latency for generating the first speech waveform can be calculated as follows:

\begin{itemize}[leftmargin=*,itemsep=0pt,parsep=0.2em,topsep=0em,partopsep=0em]

\item \textbf{Speech Tokenization:}  
   The user's speech input is processed in a streaming manner by the speech tokenizer, which operates on blocks of fixed size \( t_{\text{block}} \). Thanks to the streaming design, the tokenizer begins processing immediately and only requires the time to handle the current block, regardless of the total speech duration. Thus, the tokenization latency is:  
   \[
   T_{\text{speech\_tokenize}} = f_{\text{speech\_tokenize}}(t_{\text{block}})
   \]

\item \textbf{LLM Prefilling:}  
   The number of speech tokens, \( N_{\text{speech\_tokens}} \), generated by the tokenizer is based on the length of the user's speech $T_{\text{user\_speech}}$ and the frame rate \( fr = 12.5 \) tokens per second. The prefill latency for the LLM is given by:  
   \[
   T_{\text{llm\_prefill}} = f_{\text{llm\_prefill}}\left(fr \cdot T_{\text{user\_speech}} \right)
   \]

\item \textbf{LLM Decoding:}  
   For the initial audio response, the LLM generates 13 text tokens and 10 speech tokens, resulting in a total of \( N_{\text{first\_speech}} = 13 + 10 = 23 \) tokens. The decoding latency for this step is:  
   \[
   T_{\text{llm\_decode}} = f_{\text{llm\_decode}}\left(N_{\text{first\_speech}}\right)
   \]

\item \textbf{Speech Decoding:}  
   The \( N_{\text{speech}} = 10 \) audio tokens are processed by the speech decoder to generate the first audio chunk. The latency for this step is:  
   \[
   T_{\text{speech\_decode}} = f_{\text{speech\_decode}}\left(N_{\text{speech}}\right)
   \]

\end{itemize}

The total response latency is then:  
\[
T_{\text{total}} = T_{\text{speech\_tokenize}} + T_{\text{llm\_prefill}} + T_{\text{llm\_decode}} + T_{\text{speech\_decode}}
\]

\section{Training Procedure}

\subsection{Stage 1: Joint Speech-Text Pre-training}

We adopt the same pre-training data and procedure in \citet{speechpretrain}. The primary objective of this stage is to extend speech modeling ability to LLM through large-scale speech pre-training. We utilize three types of speech data:

\begin{itemize}[leftmargin=*,itemsep=0pt,parsep=0.2em,topsep=0.0em,partopsep=0.0em]
\item \textbf{Interleaved speech-text data:} Synthesized from text pre-training data as described in \citet{speechpretrain}, these datasets facilitate cross-modal knowledge transfer between text and speech.
\item \textbf{Unsupervised speech data:} Comprising 700k hours of speech data, this dataset encourages the model to learn from real-world speech.
\item \textbf{Supervised speech-text data:} Including both ASR and TTS data, this dataset improves the model's capabilities in basic speech tasks.
\end{itemize}

We also mix text pre-training datasets to maintain text performance. The statistics of training data is shown in~\Cref{tab:pre-training-stats}.

\subsubsection{Hyper-parameters}

\begin{wraptable}{r}{0.5\textwidth}
\centering
\vspace{-2em}
\caption{\textbf{Statistics of training data.}}
\label{tab:pre-training-stats}
\begin{tabular}{lccc}
\toprule
 & \multicolumn{2}{c}{\textbf{\# Tokens}} & \\
 & Speech & Text & \textbf{Epochs} \\ \midrule
Speech-Text & 455B & 279B  & 0.90 \\
Speech-Only & 31B & - & 2.10 \\
ASR + TTS &  11B & 3.5B & 2.07 \\
Text-only & - & 10T & 0.03 \\ \bottomrule
\end{tabular}
\vspace{-0.8em}
\end{wraptable}

We initialize \model from \texttt{GLM-4-9B-Base}~\citep{chatglm} and expand its vocabulary to include speech tokens. We perform pre-training on 1 trillion tokens, with a fixed sampling ratio of $30\%$ text data, one epoch each of unsupervised speech and supervised speech-text data, and the remainder composed of interleaved speech-text data. The composition of the training corpora is detailed in \Cref{tab:pre-training-stats}. We use the AdamW~\citep{adamw} optimizer with $\beta_1=0.9$ and $\beta_2=0.95$. The model is trained with a sequence length of 8192 and a learning rate that linearly decays from $6 \times 10^{-5}$ to $6 \times 10^{-6}$.

\subsection{Stage 2: Supervised Fine-tuning}

\subsubsection{Data Construction}

To create a human-like spoken chatbot, we utilize the following two types of data:

\begin{itemize}[leftmargin=*,itemsep=0pt,parsep=0.2em,topsep=0.0em,partopsep=0.0em]
\item \textbf{Multi-turn conversational spoken dialogues}: These dialogues are primarily derived from text-based data, carefully filtered to ensure quality. Code and math-related content are excluded to focus on conversational material suitable for spoken interactions. Responses are refined by shortening lengthy texts and avoiding outputs unsuitable for verbal delivery. Corresponding speech outputs are synthesized to align with the refined dialogues. To enhance speech input diversity in real-world voice chat scenarios, annotators read and record a variety of speech inputs.
\item \textbf{Speech style-controlled spoken dialogues}: This category contains high-quality multi-turn spoken dialogues tailored to specific speech style requirements, such as speed, emotion, or dialect.
\end{itemize}

\subsubsection{Training Details}



As described in \Cref{sec:inference}, we decouple the speech-to-speech task into two subtasks and employ the \textit{streaming thoughts} template to reduce latency. Each conversational turn consists of a user speech input $Q_s$, the corresponding text input $Q_t$, a text output $A_t$, and the corresponding speech output $A_s$. 

We observed differing learning curves for the two subtasks. Specifically, given a user speech input $Q_s$, the model learns the text output $A_t$ more quickly and compared to the speech output $A_s$. To address this discrepancy, we split each training sample into two components: one focuses on learning the text output from the speech input by masking the loss for the speech output, while the other focuses on learning the speech output from both the speech input and text output by masking the loss for the text output.

The model is fine-tuned for 20 epochs on speech output and 4 epochs on text output. The learning rate is gradually reduced from $1 \times 10^{-5}$ to $1 \times 10^{-6}$. To mitigate overfitting, we apply a weight decay of 0.1, set a dropout rate of 0.5 for hidden layers, and clip gradients to a maximum value of 1.0.

\section{Evaluation}

\subsection{Base Model Evaluation}
We evaluate the base model with two speech-text tasks, speech language modeling~\citep{AudioLM} and spoken question answering~\citep{Spectron}. For both tasks we consider two different settings: from speech context to speech generation (denoted as S$\rightarrow$S), and from speech context to text generation, denoted as S$\rightarrow$T. For all the tasks we synthesis the contexts and continuations with the multi-speaker TTS API provided by VolcEngine\footnote{\url{https://www.volcengine.com/docs/6561/79820}}.

\begin{table}[!ht]
\centering
\caption{\textbf{Speech Language Modeling results.} Results for Spirit-LM are taken from \citet{spiritlm} and other results are from \citet{moshi}.}
\label{tab:speech_lm}
\begin{tabular}{lrrrrr}
\toprule
 & \textbf{Modality} & \textbf{\# Params} & \textbf{Topic-StoryCloze} & \textbf{StoryCloze}\\ \midrule
TWIST & S$\rightarrow$S & 7B & 66.6 & 53.3 \\
Spirit-LM & S$\rightarrow$S & 7B & 82.9 & 61.0 \\
Spirit-LM & S$\rightarrow$T & 7B & 88.6 & 64.6 \\
Moshi & S$\rightarrow$S & 7B & 83.0 & 60.8 \\
\midrule
GLM-4-Voice & S$\rightarrow$T & 9B & \textbf{93.6} & \textbf{76.3} \\ 
GLM-4-Voice & S$\rightarrow$S & 9B & 82.9 & 62.4 \\ 
\bottomrule
\end{tabular}
\end{table}

\paragraph{Speech Language Modeling} This tasks evaluates the pretrained model's ability to model interleaved speech and texts. The model is given a context and required to select the correct continuation according to the predicted likelihood. We use two datasets proposed by \citet{TWIST}, spoken StoryCloze and spokeh Topic-StoryCloze. Both datasets are transformed from the the StoryCloze textual benchmark~\cite{StoryCloze}. The spoken Topic-StoryCloze is easier than spoken StoryCloze. The baseline results are taken from \citet{moshi}.

\begin{table}[t]
\centering
\caption{\textbf{Spoken Question Answering results.} Results for baselines are taken from \citet{moshi}.}
\label{tab:spoken_qa}
\begin{tabular}{lrrrrr}
\toprule
 & \textbf{Modality} & \textbf{\# Params} & \textbf{Web Questions} & \textbf{Llama Questions} & \textbf{TriviaQA} \\ \midrule
TWIST & S$\rightarrow$S & 7B & 1.5 & 4.0 & - \\
SpeechGPT & S$\rightarrow$T & 7B & 6.5 & 21.6 & 14.8 \\
Spectron & S$\rightarrow$T & 1B & 6.1 & 21.9 & - \\
Moshi & S$\rightarrow$T & 7B & 26.6 & 62.3 & 22.8 \\
Moshi & S$\rightarrow$S & 7B & 9.2 & 21.0 & 7.3 \\
\midrule
GLM-4-Voice & S$\rightarrow$T & 9B & \textbf{32.2} & \textbf{64.7} & \textbf{39.1} \\ 
GLM-4-Voice & S$\rightarrow$S & 9B & 15.9 & 50.7 & 26.5 \\ 
\bottomrule
\end{tabular}
\end{table}

\paragraph{Spoken Question Answering} Similar to closed-book question answering in NLP, spoken question answering requires the speech language model to answer spoken questions about broad factual knowledge without access to external knowledge base. We evaluate our model on 3 datasets used in \citet{moshi}, Web Questions~\cite{webquestions}, Llama Questions~\cite{Spectron}, and TriviaQA~\cite{TriviaQA}. The baseline results are taken from \citet{moshi}.

\paragraph{Results} The results for speech language modeling are shown in \Cref{tab:speech_lm} and those for spoken question answering are shown in \Cref{tab:spoken_qa}. We can observe that \model outperforms baselines on all the evaluated tasks in both S$\rightarrow$S and S$\rightarrow$T settings, except Topic-StoryCloze in the S$\rightarrow$S setting. Compared with Moshi~\cite{moshi}, which also supports both speech and text modalities, our model excels in spoken question answering, whether the answers are textual or spoken. Another observation is that the accuracy in the S$\rightarrow$T setting is always better than that in the S$\rightarrow$S setting, especially for spoken question answering. Therefore textual guidance is still necessary for intelligent speech chatbots. However, our method significantly reduces the gap between spoken answers and textual answers on spoken question answering, especially on Llama Questions, with the potential to develop direct speech-to-speech chatbots.

\paragraph{ASR / TTS} We prompt the base model with the same prompt format used for the ASR / TTS task in pre-training. Whisper-Large-V3~\citep{whisper} and Paraformer-Large~\citep{shi2023seaco} are employed to generate the text prediction for English and Chinese recognition in the TTS task respectively. Before computing the error rate, the text prediction is normalized respectively with tokenizer of \texttt{whisper-large-v3} and CosyVoice~\citep{cosyvoice} pipeline for ASR and TTS tasks. The results are summarized in Table~\ref{tab:model_asr_tts}. GLM-4-Voice achieve similar ASR and TTS ability compared with \texttt{whisper-large-v3}\citep{whisper} and CosyVoice~\citep{cosyvoice} baselines.

\begin{table}[t]
    \centering
    \caption{\textbf{ASR and TTS results}. The LibriSpeech (English) is measured with word-error-rate (WER) and AISHELL-1 (Chinese) is measured with character-error-rate (CER). The TTS tasks are measured with WER. We use $\emptyset$ to indicate tasks and modalities not supported by the model.}
    \begin{tabular}{lccc|ccc}
    \toprule
     & \multicolumn{2}{c}{\textbf{LibriSpeech}} & \textbf{AISHELL-1} & \textbf{LibriTTS} & \multicolumn{2}{c}{\textbf{Seed-TTS}} \\
     & \multicolumn{1}{l}{test-clean} & \multicolumn{1}{l}{test-other} & \multicolumn{1}{c|}{test} & \multicolumn{1}{l}{test-clean} & \multicolumn{1}{l}{test-en} & \multicolumn{1}{l}{test-zh} \\ \midrule
    CosyVoice &  $\emptyset$ & $\emptyset$ & $\emptyset$ & \textbf{3.17} & 3.39 & 3.10 \\
    whisper-large-v3 & \textbf{2.50} & \textbf{4.53} & 9.31 & $\emptyset$ & $\emptyset$ & $\emptyset$ \\ \midrule
    GLM-4-Voice & 2.82 & 7.66 & \textbf{2.46} & 5.64 & \textbf{2.91} & \textbf{2.10}  \\ \bottomrule
    \end{tabular}
    \label{tab:model_asr_tts}
\end{table}

\subsection{Chat Model Evaluation}

\paragraph{ChatGPT Score} To evaluate the question answering ability and knowledge memorization
of the fine-tuned chat model, we use GPT-4o~\citep{gpt4o}, specifically \texttt{gpt-4o-2024-05-13}, to evaluate quality or correctness of the model response. 
For the General QA task, we adopt the questions from the \texttt{helpful base} and \texttt{vicuna} subset of AlpacaEval~\citep{alpacaeval} with math-related questions removed, which follows the chat evaluation dateset of Llama-Omni~\citep{llama-omni}.
We ask GPT-4o to evaluate response quality and score the response in a range from 1 to 10 following the evaluation method of MT-Bench~\citep{zheng2023judging}.
For the Knowledge task, we select 100 questions from Web Questions, Llama Questions, and TriviaQA. We provide GPT-4o with ground-truth answer and ask it to judge whether the response of the model is correct.
The score reported in Table~\ref{tab:chat_model_results} is the answer accuracy normalized to a scale of 0 (0\%) to 10 (100\%).
All texts used for judging are audio transcriptions produced by Whisper-Large-V3~\citep{whisper} and the prompts used for scoring are included in Appendix~\ref{app:prompt-for-evaluation}.

\paragraph{Speech Quality} We use the UTMOS~\citep{saeki2022utmos} model to predict the mean opinion score (MOS) to evaluate the naturalness of the generated speech.

\paragraph{Speech-Text Alignment} To evaluate the correspondence between the generated text responses and speech responses, we transcribe the speech responses for the General QA task into text with \textit{whipser-large-v3}~\citep{whisper}. Then, the word error rate (WER) is calculated between the transcription and the text response, which is referred to as ASR-WER(\%) in Table~\ref{tab:chat_model_results}. \model is a bilingual model and sometimes answers the English query with a Chinese response, whose WER cannot be calculated directly. For a fair comparison with the English-only baseline models, we restrict the output of \model to English tokens when evaluating the tasks reported in Table~\ref{tab:chat_model_results}.

\begin{table}[t]
\centering
\caption{\textbf{Chat model evaluation results.} The baseline results are taken from \citet{speechpretrain}}
\begin{tabular}{lcccc}
\toprule
  & \multicolumn{2}{c}{\textbf{ChatGPT Score ↑}} & \multirow{2}{*}{\textbf{UTMOS ↑}} & \multirow{2}{*}{\textbf{ASR-WER ↓}} \\ 
\multicolumn{1}{l}{} & \multicolumn{1}{c}{General QA} & \multicolumn{1}{c}{Knowledge} &  &  \\ \midrule
    SpeechGPT~\citep{speechgpt}    & 1.40 & 2.20 &  3.86 & 66.57 \\ 
    Mini-Omni~\citep{mini-omni}    & 2.44 & 1.10 & 3.17 & 25.28 \\ 
    Llama-Omni~\citep{llama-omni} & 3.50 & 3.90 & 3.92 & 9.18 \\ 
    Moshi~\citep{moshi}  & 2.42 & 3.60 & 3.90 & 7.95 \\
    \midrule
    GLM-4-Voice & \textbf{5.40} & \textbf{5.20} & \textbf{4.45} & \textbf{5.74} \\
\bottomrule
\end{tabular}
\label{tab:chat_model_results}
\end{table}

\section{Conclusion}

In this paper, we introduced \model, an end-to-end spoken chatbot designed for natural and expressive voice interactions. By integrating a 12.5Hz supverised speech tokenizer, a flow-matching based speech decoder, and large-scale pre-training on 1 trillion tokens of speech-text data, \model effectively bridges text and speech modalities. It achieves strong performance across tasks like speech language modeling, ASR, TTS, and spoken question answering. Fine-tuning with high-quality conversational datasets further enhances its ability to generate fluent, low-latency, and nuanced responses. The open availability of \model encourages further exploration in building practical and accessible spoken AI systems.

\clearpage

\bibliographystyle{plainnat}
\bibliography{ref}

\begin{thebibliography}{49}
\providecommand{\natexlab}[1]{#1}
\providecommand{\url}[1]{\texttt{#1}}
\expandafter\ifx\csname urlstyle\endcsname\relax
  \providecommand{\doi}[1]{doi: #1}\else
  \providecommand{\doi}{doi: \begingroup \urlstyle{rm}\Url}\fi

\bibitem[An et~al.(2024)An, Chen, Deng, Du, Gao, Gao, Gu, He, Hu, Hu, Ji, Li, Li, Lu, Luo, Lv, Ma, Ma, Ni, Song, Shi, Shi, Wang, Wang, Wang, Xiao, Yan, Yang, Zhang, Zhang, Zhang, Zhao, and Zheng]{funasr}
Keyu An, Qian Chen, Chong Deng, Zhihao Du, Changfeng Gao, Zhifu Gao, Yue Gu, Ting He, Hangrui Hu, Kai Hu, Shengpeng Ji, Yabin Li, Zerui Li, Heng Lu, Haoneng Luo, Xiang Lv, Bin Ma, Ziyang Ma, Chongjia Ni, Changhe Song, Jiaqi Shi, Xian Shi, Hao Wang, Wen Wang, Yuxuan Wang, Zhangyu Xiao, Zhijie Yan, Yexin Yang, Bin Zhang, Qinglin Zhang, Shiliang Zhang, Nan Zhao, and Siqi Zheng.
\newblock Funaudiollm: Voice understanding and generation foundation models for natural interaction between humans and llms.
\newblock \emph{CoRR}, abs/2407.04051, 2024.
\newblock URL \url{https://doi.org/10.48550/arXiv.2407.04051}.

\bibitem[Ao et~al.(2022)Ao, Wang, Zhou, Wang, Ren, Wu, Liu, Ko, Li, Zhang, et~al.]{speecht5}
Junyi Ao, Rui Wang, Long Zhou, Chengyi Wang, Shuo Ren, Yu~Wu, Shujie Liu, Tom Ko, Qing Li, Yu~Zhang, et~al.
\newblock Speecht5: Unified-modal encoder-decoder pre-training for spoken language processing.
\newblock In \emph{Proceedings of the 60th Annual Meeting of the Association for Computational Linguistics (Volume 1: Long Papers)}, pages 5723--5738, 2022.

\bibitem[Ardila et~al.(2020)Ardila, Branson, Davis, Kohler, Meyer, Henretty, Morais, Saunders, Tyers, and Weber]{commonvoice}
Rosana Ardila, Megan Branson, Kelly Davis, Michael Kohler, Josh Meyer, Michael Henretty, Reuben Morais, Lindsay Saunders, Francis~M. Tyers, and Gregor Weber.
\newblock Common voice: {A} massively-multilingual speech corpus.
\newblock In \emph{Proceedings of The 12th Language Resources and Evaluation Conference, {LREC} 2020, Marseille, France, May 11-16, 2020}, pages 4218--4222. European Language Resources Association, 2020.

\bibitem[Berant et~al.(2013)Berant, Chou, Frostig, and Liang]{webquestions}
Jonathan Berant, Andrew Chou, Roy Frostig, and Percy Liang.
\newblock Semantic parsing on freebase from question-answer pairs.
\newblock In \emph{Proceedings of the 2013 Conference on Empirical Methods in Natural Language Processing, {EMNLP} 2013, 18-21 October 2013, Grand Hyatt Seattle, Seattle, Washington, USA, {A} meeting of SIGDAT, a Special Interest Group of the {ACL}}, pages 1533--1544. {ACL}, 2013.

\bibitem[Borsos et~al.(2023)Borsos, Marinier, Vincent, Kharitonov, Pietquin, Sharifi, Roblek, Teboul, Grangier, Tagliasacchi, and Zeghidour]{AudioLM}
Zal{\'{a}}n Borsos, Rapha{\"{e}}l Marinier, Damien Vincent, Eugene Kharitonov, Olivier Pietquin, Matthew Sharifi, Dominik Roblek, Olivier Teboul, David Grangier, Marco Tagliasacchi, and Neil Zeghidour.
\newblock Audiolm: {A} language modeling approach to audio generation.
\newblock \emph{{IEEE} {ACM} Trans. Audio Speech Lang. Process.}, 31:\penalty0 2523--2533, 2023.

\bibitem[Bu et~al.(2017)Bu, Du, Na, Wu, and Zheng]{aishell1}
Hui Bu, Jiayu Du, Xingyu Na, Bengu Wu, and Hao Zheng.
\newblock {AISHELL-1:} an open-source mandarin speech corpus and a speech recognition baseline.
\newblock In \emph{20th Conference of the Oriental Chapter of the International Coordinating Committee on Speech Databases and Speech {I/O} Systems and Assessment, {O-COCOSDA} 2017, Seoul, South Korea, November 1-3, 2017}, pages 1--5. {IEEE}, 2017.

\bibitem[Chen et~al.(2021{\natexlab{a}})Chen, Chai, Wang, Du, Zhang, Weng, Su, Povey, Trmal, Zhang, Jin, Khudanpur, Watanabe, Zhao, Zou, Li, Yao, Wang, You, and Yan]{GigaSpeech}
Guoguo Chen, Shuzhou Chai, Guan{-}Bo Wang, Jiayu Du, Wei{-}Qiang Zhang, Chao Weng, Dan Su, Daniel Povey, Jan Trmal, Junbo Zhang, Mingjie Jin, Sanjeev Khudanpur, Shinji Watanabe, Shuaijiang Zhao, Wei Zou, Xiangang Li, Xuchen Yao, Yongqing Wang, Zhao You, and Zhiyong Yan.
\newblock Gigaspeech: An evolving, multi-domain {ASR} corpus with 10, 000 hours of transcribed audio.
\newblock In \emph{22nd Annual Conference of the International Speech Communication Association, Interspeech 2021, Brno, Czechia, August 30 - September 3, 2021}, pages 3670--3674. {ISCA}, 2021{\natexlab{a}}.

\bibitem[Chen et~al.(2021{\natexlab{b}})Chen, Chi, Yang, Chang, Lin, Huang, Liu, Liu, Lee, and Lee]{speechnet}
Yi-Chen Chen, Po-Han Chi, Shu-wen Yang, Kai-Wei Chang, Jheng-hao Lin, Sung-Feng Huang, Da-Rong Liu, Chi-Liang Liu, Cheng-Kuang Lee, and Hung-yi Lee.
\newblock Speechnet: A universal modularized model for speech processing tasks.
\newblock \emph{arXiv preprint arXiv:2105.03070}, 2021{\natexlab{b}}.

\bibitem[Chu et~al.(2023)Chu, Xu, Zhou, Yang, Zhang, Yan, Zhou, and Zhou]{qwen-audio}
Yunfei Chu, Jin Xu, Xiaohuan Zhou, Qian Yang, Shiliang Zhang, Zhijie Yan, Chang Zhou, and Jingren Zhou.
\newblock Qwen-audio: Advancing universal audio understanding via unified large-scale audio-language models.
\newblock \emph{CoRR}, abs/2311.07919, 2023.

\bibitem[Chung et~al.(2021)Chung, Zhang, Han, Chiu, Qin, Pang, and Wu]{w2vbert}
Yu{-}An Chung, Yu~Zhang, Wei Han, Chung{-}Cheng Chiu, James Qin, Ruoming Pang, and Yonghui Wu.
\newblock w2v-bert: Combining contrastive learning and masked language modeling for self-supervised speech pre-training.
\newblock In \emph{{IEEE} Automatic Speech Recognition and Understanding Workshop, {ASRU} 2021, Cartagena, Colombia, December 13-17, 2021}, pages 244--250. {IEEE}, 2021.

\bibitem[D{\'{e}}fossez et~al.(2023)D{\'{e}}fossez, Copet, Synnaeve, and Adi]{neuralac}
Alexandre D{\'{e}}fossez, Jade Copet, Gabriel Synnaeve, and Yossi Adi.
\newblock High fidelity neural audio compression.
\newblock \emph{Trans. Mach. Learn. Res.}, 2023, 2023.

\bibitem[D\'efossez et~al.(2024)D\'efossez, Mazar\'e, Orsini, Royer, P\'erez, J\'egou, Grave, and Zeghidour]{moshi}
Alexandre D\'efossez, Laurent Mazar\'e, Manu Orsini, Am\'elie Royer, Patrick P\'erez, Herv\'e J\'egou, Edouard Grave, and Neil Zeghidour.
\newblock Moshi: a speech-text foundation model for real-time dialogue.
\newblock Technical report, Kyutai, September 2024.
\newblock URL \url{http://kyutai.org/Moshi.pdf}.

\bibitem[Dhariwal et~al.(2020)Dhariwal, Jun, Payne, Kim, Radford, and Sutskever]{jukebox}
Prafulla Dhariwal, Heewoo Jun, Christine Payne, Jong~Wook Kim, Alec Radford, and Ilya Sutskever.
\newblock Jukebox: {A} generative model for music.
\newblock \emph{CoRR}, abs/2005.00341, 2020.

\bibitem[Du et~al.(2024)Du, Chen, Zhang, Hu, Lu, Yang, Hu, Zheng, Gu, Ma, Gao, and Yan]{cosyvoice}
Zhihao Du, Qian Chen, Shiliang Zhang, Kai Hu, Heng Lu, Yexin Yang, Hangrui Hu, Siqi Zheng, Yue Gu, Ziyang Ma, Zhifu Gao, and Zhijie Yan.
\newblock Cosyvoice: A scalable multilingual zero-shot text-to-speech synthesizer based on supervised semantic tokens, 2024.
\newblock URL \url{https://arxiv.org/abs/2407.05407}.

\bibitem[Fang et~al.(2024)Fang, Guo, Zhou, Ma, Zhang, and Feng]{llama-omni}
Qingkai Fang, Shoutao Guo, Yan Zhou, Zhengrui Ma, Shaolei Zhang, and Yang Feng.
\newblock Llama-omni: Seamless speech interaction with large language models, 2024.
\newblock URL \url{https://arxiv.org/abs/2409.06666}.

\bibitem[GLM et~al.(2024)GLM, Zeng, Xu, Wang, Zhang, Yin, Zhang, Rojas, Feng, Zhao, Lai, Yu, Wang, Sun, Zhang, Cheng, Gui, Tang, Zhang, Sun, Li, Zhao, Wu, Zhong, Liu, Huang, Zhang, Zheng, Lu, Duan, Zhang, Cao, Yang, Tam, Zhao, Liu, Xia, Zhang, Gu, Lv, Liu, Liu, Yang, Song, Zhang, An, Xu, Niu, Yang, Li, Bai, Dong, Qi, Wang, Yang, Du, Hou, and Wang]{chatglm}
Team GLM, Aohan Zeng, Bin Xu, Bowen Wang, Chenhui Zhang, Da~Yin, Dan Zhang, Diego Rojas, Guanyu Feng, Hanlin Zhao, Hanyu Lai, Hao Yu, Hongning Wang, Jiadai Sun, Jiajie Zhang, Jiale Cheng, Jiayi Gui, Jie Tang, Jing Zhang, Jingyu Sun, Juanzi Li, Lei Zhao, Lindong Wu, Lucen Zhong, Mingdao Liu, Minlie Huang, Peng Zhang, Qinkai Zheng, Rui Lu, Shuaiqi Duan, Shudan Zhang, Shulin Cao, Shuxun Yang, Weng~Lam Tam, Wenyi Zhao, Xiao Liu, Xiao Xia, Xiaohan Zhang, Xiaotao Gu, Xin Lv, Xinghan Liu, Xinyi Liu, Xinyue Yang, Xixuan Song, Xunkai Zhang, Yifan An, Yifan Xu, Yilin Niu, Yuantao Yang, Yueyan Li, Yushi Bai, Yuxiao Dong, Zehan Qi, Zhaoyu Wang, Zhen Yang, Zhengxiao Du, Zhenyu Hou, and Zihan Wang.
\newblock Chatglm: A family of large language models from glm-130b to glm-4 all tools, 2024.
\newblock URL \url{https://arxiv.org/abs/2406.12793}.

\bibitem[Hassid et~al.(2023)Hassid, Remez, Nguyen, Gat, Conneau, Kreuk, Copet, D{\'{e}}fossez, Synnaeve, Dupoux, Schwartz, and Adi]{TWIST}
Michael Hassid, Tal Remez, Tu~Anh Nguyen, Itai Gat, Alexis Conneau, Felix Kreuk, Jade Copet, Alexandre D{\'{e}}fossez, Gabriel Synnaeve, Emmanuel Dupoux, Roy Schwartz, and Yossi Adi.
\newblock Textually pretrained speech language models.
\newblock In \emph{Advances in Neural Information Processing Systems 36: Annual Conference on Neural Information Processing Systems 2023, NeurIPS 2023, New Orleans, LA, USA, December 10 - 16, 2023}, 2023.

\bibitem[Hines et~al.(2015)Hines, Skoglund, Kokaram, and Harte]{visqol}
Andrew Hines, Jan Skoglund, Anil Kokaram, and Naomi Harte.
\newblock Visqol: an objective speech quality model.
\newblock \emph{EURASIP Journal on Audio, Speech, and Music Processing}, 2015 (13):\penalty0 1--18, 2015.

\bibitem[Hsu et~al.(2021)Hsu, Bolte, Tsai, Lakhotia, Salakhutdinov, and Mohamed]{hubert}
Wei{-}Ning Hsu, Benjamin Bolte, Yao{-}Hung~Hubert Tsai, Kushal Lakhotia, Ruslan Salakhutdinov, and Abdelrahman Mohamed.
\newblock Hubert: Self-supervised speech representation learning by masked prediction of hidden units.
\newblock \emph{{IEEE} {ACM} Trans. Audio Speech Lang. Process.}, 29:\penalty0 3451--3460, 2021.

\bibitem[Ji et~al.(2024)Ji, Jiang, Cheng, Chen, Fang, Zuo, Yang, Li, Zhang, Yang, Huang, Jiang, Chen, Zheng, Wang, and Zhao]{wavtokenizer}
Shengpeng Ji, Ziyue Jiang, Xize Cheng, Yifu Chen, Minghui Fang, Jialong Zuo, Qian Yang, Ruiqi Li, Ziang Zhang, Xiaoda Yang, Rongjie Huang, Yidi Jiang, Qian Chen, Siqi Zheng, Wen Wang, and Zhou Zhao.
\newblock Wavtokenizer: an efficient acoustic discrete codec tokenizer for audio language modeling.
\newblock \emph{CoRR}, abs/2408.16532, 2024.

\bibitem[Joshi et~al.(2017)Joshi, Choi, Weld, and Zettlemoyer]{TriviaQA}
Mandar Joshi, Eunsol Choi, Daniel~S. Weld, and Luke Zettlemoyer.
\newblock Triviaqa: {A} large scale distantly supervised challenge dataset for reading comprehension.
\newblock In \emph{Proceedings of the 55th Annual Meeting of the Association for Computational Linguistics, {ACL} 2017, Vancouver, Canada, July 30 - August 4, Volume 1: Long Papers}, pages 1601--1611. Association for Computational Linguistics, 2017.

\bibitem[Kong et~al.(2020)Kong, Kim, and Bae]{hifi_gan}
Jungil Kong, Jaehyeon Kim, and Jaekyoung Bae.
\newblock Hifi-gan: Generative adversarial networks for efficient and high fidelity speech synthesis.
\newblock In H.~Larochelle, M.~Ranzato, R.~Hadsell, M.F. Balcan, and H.~Lin, editors, \emph{Advances in Neural Information Processing Systems}, volume~33, pages 17022--17033. Curran Associates, Inc., 2020.
\newblock URL \url{https://proceedings.neurips.cc/paper_files/paper/2020/file/c5d736809766d46260d816d8dbc9eb44-Paper.pdf}.

\bibitem[Kumar et~al.(2023)Kumar, Seetharaman, Luebs, Kumar, and Kumar]{audio_rvqgan}
Rithesh Kumar, Prem Seetharaman, Alejandro Luebs, Ishaan Kumar, and Kundan Kumar.
\newblock High-fidelity audio compression with improved {RVQGAN}.
\newblock In \emph{Advances in Neural Information Processing Systems 36: Annual Conference on Neural Information Processing Systems 2023, NeurIPS 2023, New Orleans, LA, USA, December 10 - 16, 2023}, 2023.

\bibitem[Lakhotia et~al.(2021)Lakhotia, Kharitonov, Hsu, Adi, Polyak, Bolte, Nguyen, Copet, Baevski, Mohamed, and Dupoux]{GSLM}
Kushal Lakhotia, Eugene Kharitonov, Wei-Ning Hsu, Yossi Adi, Adam Polyak, Benjamin Bolte, Tu-Anh Nguyen, Jade Copet, Alexei Baevski, Abdelrahman Mohamed, and Emmanuel Dupoux.
\newblock On generative spoken language modeling from raw audio.
\newblock \emph{Transactions of the Association for Computational Linguistics}, 9:\penalty0 1336--1354, 2021.

\bibitem[Li et~al.(2023)Li, Zhang, Dubois, Taori, Gulrajani, Guestrin, Liang, and Hashimoto]{alpacaeval}
Xuechen Li, Tianyi Zhang, Yann Dubois, Rohan Taori, Ishaan Gulrajani, Carlos Guestrin, Percy Liang, and Tatsunori~B. Hashimoto.
\newblock Alpacaeval: An automatic evaluator of instruction-following models.
\newblock \url{https://github.com/tatsu-lab/alpaca_eval}, 5 2023.

\bibitem[Lo et~al.(2019)Lo, Fu, Huang, Wang, Yamagishi, Tsao, and Wang]{MOSNet}
Chen{-}Chou Lo, Szu{-}Wei Fu, Wen{-}Chin Huang, Xin Wang, Junichi Yamagishi, Yu~Tsao, and Hsin{-}Min Wang.
\newblock Mosnet: Deep learning-based objective assessment for voice conversion.
\newblock In Gernot Kubin and Zdravko Kacic, editors, \emph{20th Annual Conference of the International Speech Communication Association, Interspeech 2019, Graz, Austria, September 15-19, 2019}, pages 1541--1545. {ISCA}, 2019.
\newblock \doi{10.21437/INTERSPEECH.2019-2003}.
\newblock URL \url{https://doi.org/10.21437/Interspeech.2019-2003}.

\bibitem[Loshchilov and Hutter(2019)]{adamw}
Ilya Loshchilov and Frank Hutter.
\newblock Decoupled weight decay regularization, 2019.
\newblock URL \url{https://arxiv.org/abs/1711.05101}.

\bibitem[Mehta et~al.(2024)Mehta, Tu, Beskow, Sz{\'e}kely, and Henter]{mehta2024matcha}
Shivam Mehta, Ruibo Tu, Jonas Beskow, {\'E}va Sz{\'e}kely, and Gustav~Eje Henter.
\newblock Matcha-{TTS}: A fast {TTS} architecture with conditional flow matching.
\newblock In \emph{Proc. ICASSP}, 2024.

\bibitem[Mostafazadeh et~al.(2016)Mostafazadeh, Chambers, He, Parikh, Batra, Vanderwende, Kohli, and Allen]{StoryCloze}
Nasrin Mostafazadeh, Nathanael Chambers, Xiaodong He, Devi Parikh, Dhruv Batra, Lucy Vanderwende, Pushmeet Kohli, and James~F. Allen.
\newblock A corpus and evaluation framework for deeper understanding of commonsense stories.
\newblock \emph{CoRR}, abs/1604.01696, 2016.

\bibitem[Nachmani et~al.(2024)Nachmani, Levkovitch, Hirsch, Salazar, Asawaroengchai, Mariooryad, Rivlin, Skerry{-}Ryan, and Ramanovich]{Spectron}
Eliya Nachmani, Alon Levkovitch, Roy Hirsch, Julian Salazar, Chulayuth Asawaroengchai, Soroosh Mariooryad, Ehud Rivlin, R.~J. Skerry{-}Ryan, and Michelle~Tadmor Ramanovich.
\newblock Spoken question answering and speech continuation using spectrogram-powered {LLM}.
\newblock In \emph{The Twelfth International Conference on Learning Representations, {ICLR} 2024, Vienna, Austria, May 7-11, 2024}. OpenReview.net, 2024.

\bibitem[Nguyen et~al.(2023)Nguyen, Hsu, D'Avirro, Shi, Gat, Fazel{-}Zarandi, Remez, Copet, Synnaeve, Hassid, Kreuk, Adi, and Dupoux]{Expresso}
Tu~Anh Nguyen, Wei{-}Ning Hsu, Antony D'Avirro, Bowen Shi, Itai Gat, Maryam Fazel{-}Zarandi, Tal Remez, Jade Copet, Gabriel Synnaeve, Michael Hassid, Felix Kreuk, Yossi Adi, and Emmanuel Dupoux.
\newblock Expresso: {A} benchmark and analysis of discrete expressive speech resynthesis.
\newblock In Naomi Harte, Julie Carson{-}Berndsen, and Gareth Jones, editors, \emph{24th Annual Conference of the International Speech Communication Association, Interspeech 2023, Dublin, Ireland, August 20-24, 2023}, pages 4823--4827. {ISCA}, 2023.

\bibitem[Nguyen et~al.(2024)Nguyen, Muller, Yu, Costa-jussa, Elbayad, Popuri, Duquenne, Algayres, Mavlyutov, Gat, Synnaeve, Pino, Sagot, and Dupoux]{spiritlm}
Tu~Anh Nguyen, Benjamin Muller, Bokai Yu, Marta~R. Costa-jussa, Maha Elbayad, Sravya Popuri, Paul-Ambroise Duquenne, Robin Algayres, Ruslan Mavlyutov, Itai Gat, Gabriel Synnaeve, Juan Pino, Benoit Sagot, and Emmanuel Dupoux.
\newblock Spirit-lm: Interleaved spoken and written language model, 2024.
\newblock URL \url{https://arxiv.org/abs/2402.05755}.

\bibitem[OpenAI(2024)]{gpt4o}
OpenAI.
\newblock Hello gpt-4o, 2024.
\newblock URL \url{https://openai.com/index/hello-gpt-4o/}.

\bibitem[Panayotov et~al.(2015)Panayotov, Chen, Povey, and Khudanpur]{librispeech}
Vassil Panayotov, Guoguo Chen, Daniel Povey, and Sanjeev Khudanpur.
\newblock Librispeech: An asr corpus based on public domain audio books.
\newblock In \emph{2015 IEEE International Conference on Acoustics, Speech and Signal Processing (ICASSP)}, pages 5206--5210, 2015.
\newblock \doi{10.1109/ICASSP.2015.7178964}.

\bibitem[Pratap et~al.(2020)Pratap, Xu, Sriram, Synnaeve, and Collobert]{MLS}
Vineel Pratap, Qiantong Xu, Anuroop Sriram, Gabriel Synnaeve, and Ronan Collobert.
\newblock {MLS:} {A} large-scale multilingual dataset for speech research.
\newblock In \emph{21st Annual Conference of the International Speech Communication Association, Interspeech 2020, Virtual Event, Shanghai, China, October 25-29, 2020}, pages 2757--2761. {ISCA}, 2020.

\bibitem[Radford et~al.(2023)Radford, Kim, Xu, Brockman, McLeavey, and Sutskever]{whisper}
Alec Radford, Jong~Wook Kim, Tao Xu, Greg Brockman, Christine McLeavey, and Ilya Sutskever.
\newblock Robust speech recognition via large-scale weak supervision.
\newblock In Andreas Krause, Emma Brunskill, Kyunghyun Cho, Barbara Engelhardt, Sivan Sabato, and Jonathan Scarlett, editors, \emph{International Conference on Machine Learning, {ICML} 2023, 23-29 July 2023, Honolulu, Hawaii, {USA}}, volume 202 of \emph{Proceedings of Machine Learning Research}, pages 28492--28518. {PMLR}, 2023.

\bibitem[Saeki et~al.(2022)Saeki, Xin, Nakata, Koriyama, Takamichi, and Saruwatari]{saeki2022utmos}
Takaaki Saeki, Detai Xin, Wataru Nakata, Tomoki Koriyama, Shinnosuke Takamichi, and Hiroshi Saruwatari.
\newblock Utmos: Utokyo-sarulab system for voicemos challenge 2022.
\newblock \emph{Interspeech 2022}, 2022.

\bibitem[Shi et~al.(2023)Shi, Yang, Li, and Zhang]{shi2023seaco}
Xian Shi, Yexin Yang, Zerui Li, and Shiliang Zhang.
\newblock Seaco-paraformer: A non-autoregressive asr system with flexible and effective hotword customization ability.
\newblock \emph{arXiv preprint arXiv:2308.03266 (accepted by ICASSP2024)}, 2023.

\bibitem[van~den Oord et~al.(2016)van~den Oord, Dieleman, Zen, Simonyan, Vinyals, Graves, Kalchbrenner, Senior, and Kavukcuoglu]{wavenet}
A{\"{a}}ron van~den Oord, Sander Dieleman, Heiga Zen, Karen Simonyan, Oriol Vinyals, Alex Graves, Nal Kalchbrenner, Andrew~W. Senior, and Koray Kavukcuoglu.
\newblock Wavenet: {A} generative model for raw audio.
\newblock In \emph{The 9th {ISCA} Speech Synthesis Workshop, {SSW} 2016, Sunnyvale, CA, USA, September 13-15, 2016}, page 125. {ISCA}, 2016.

\bibitem[van~den Oord et~al.(2017)van~den Oord, Vinyals, and Kavukcuoglu]{vqvae}
A{\"{a}}ron van~den Oord, Oriol Vinyals, and Koray Kavukcuoglu.
\newblock Neural discrete representation learning.
\newblock In Isabelle Guyon, Ulrike von Luxburg, Samy Bengio, Hanna~M. Wallach, Rob Fergus, S.~V.~N. Vishwanathan, and Roman Garnett, editors, \emph{Advances in Neural Information Processing Systems 30: Annual Conference on Neural Information Processing Systems 2017, December 4-9, 2017, Long Beach, CA, {USA}}, pages 6306--6315, 2017.

\bibitem[Wang et~al.(2024)Wang, Li, Fu, Xie, Li, Sun, and Ma]{freezeomni}
Xiong Wang, Yangze Li, Chaoyou Fu, Lei Xie, Ke~Li, Xing Sun, and Long Ma.
\newblock Freeze-omni: A smart and low latency speech-to-speech dialogue model with frozen llm, 2024.
\newblock URL \url{https://arxiv.org/abs/2411.00774}.

\bibitem[Xie and Wu(2024)]{mini-omni}
Zhifei Xie and Changqiao Wu.
\newblock Mini-omni: Language models can hear, talk while thinking in streaming, 2024.
\newblock URL \url{https://arxiv.org/abs/2408.16725}.

\bibitem[Yao et~al.(2021)Yao, Wu, Wang, Zhang, Yu, Yang, Peng, Chen, Xie, and Lei]{wenet}
Zhuoyuan Yao, Di~Wu, Xiong Wang, Binbin Zhang, Fan Yu, Chao Yang, Zhendong Peng, Xiaoyu Chen, Lei Xie, and Xin Lei.
\newblock Wenet: Production oriented streaming and non-streaming end-to-end speech recognition toolkit.
\newblock In \emph{22nd Annual Conference of the International Speech Communication Association, Interspeech 2021, Brno, Czechia, August 30 - September 3, 2021}, pages 4054--4058. {ISCA}, 2021.

\bibitem[Zeghidour et~al.(2022)Zeghidour, Luebs, Omran, Skoglund, and Tagliasacchi]{SoundStream}
Neil Zeghidour, Alejandro Luebs, Ahmed Omran, Jan Skoglund, and Marco Tagliasacchi.
\newblock Soundstream: An end-to-end neural audio codec.
\newblock \emph{{IEEE} {ACM} Trans. Audio Speech Lang. Process.}, 30:\penalty0 495--507, 2022.
\newblock \doi{10.1109/TASLP.2021.3129994}.
\newblock URL \url{https://doi.org/10.1109/TASLP.2021.3129994}.

\bibitem[Zeng et~al.(2024)Zeng, Du, Liu, Zhang, Jiang, Dong, and Tang]{speechpretrain}
Aohan Zeng, Zhengxiao Du, Mingdao Liu, Lei Zhang, Shengmin Jiang, Yuxiao Dong, and Jie Tang.
\newblock Scaling speech-text pre-training with synthetic interleaved data, 2024.
\newblock URL \url{https://arxiv.org/abs/2411.17607}.

\bibitem[Zhang et~al.(2023)Zhang, Li, Zhang, Zhan, Wang, Zhou, and Qiu]{speechgpt}
Dong Zhang, Shimin Li, Xin Zhang, Jun Zhan, Pengyu Wang, Yaqian Zhou, and Xipeng Qiu.
\newblock Speechgpt: Empowering large language models with intrinsic cross-modal conversational abilities, 2023.
\newblock URL \url{https://arxiv.org/abs/2305.11000}.

\bibitem[Zhang et~al.(2022)Zhang, Roller, Goyal, Artetxe, Chen, Chen, Dewan, Diab, Li, Lin, Mihaylov, Ott, Shleifer, Shuster, Simig, Koura, Sridhar, Wang, and Zettlemoyer]{OPT}
Susan Zhang, Stephen Roller, Naman Goyal, Mikel Artetxe, Moya Chen, Shuohui Chen, Christopher Dewan, Mona~T. Diab, Xian Li, Xi~Victoria Lin, Todor Mihaylov, Myle Ott, Sam Shleifer, Kurt Shuster, Daniel Simig, Punit~Singh Koura, Anjali Sridhar, Tianlu Wang, and Luke Zettlemoyer.
\newblock {OPT:} open pre-trained transformer language models.
\newblock \emph{CoRR}, abs/2205.01068, 2022.

\bibitem[Zhang et~al.(2024)Zhang, Zhang, Li, Zhou, and Qiu]{speechtokenizer}
Xin Zhang, Dong Zhang, Shimin Li, Yaqian Zhou, and Xipeng Qiu.
\newblock Speechtokenizer: Unified speech tokenizer for speech language models.
\newblock In \emph{The Twelfth International Conference on Learning Representations, {ICLR} 2024, Vienna, Austria, May 7-11, 2024}. OpenReview.net, 2024.

\bibitem[Zheng et~al.(2023)Zheng, Chiang, Sheng, Zhuang, Wu, Zhuang, Lin, Li, Li, Xing, Zhang, Gonzalez, and Stoica]{zheng2023judging}
Lianmin Zheng, Wei-Lin Chiang, Ying Sheng, Siyuan Zhuang, Zhanghao Wu, Yonghao Zhuang, Zi~Lin, Zhuohan Li, Dacheng Li, Eric~P. Xing, Hao Zhang, Joseph~E. Gonzalez, and Ion Stoica.
\newblock Judging llm-as-a-judge with mt-bench and chatbot arena, 2023.
\newblock URL \url{https://arxiv.org/abs/2306.05685}.

\end{thebibliography}

\clearpage
\appendix
\section{Appendix}

\subsection{Prompt for Evaluating Spoken Chatbots}
\label{app:prompt-for-evaluation}

\begin{tcolorbox}[left=0mm,right=0mm,top=0mm,bottom=0mm,boxsep=1mm,arc=0mm,boxrule=0pt, frame empty, breakable]
\textbf{General QA}
\small
\begin{lstlisting}
[Instruction]
Please act as an impartial judge and evaluate the quality of the response provided by an AI assistant to the user question displayed below. Your evaluation should consider factors such as the helpfulness, relevance, accuracy, depth, creativity, and level of detail of the response. Begin your evaluation by providing a short explanation. Be as objective as possible. After providing your explanation, you must rate the response on a scale of 1 to 10 by strictly following this format: "[[rating]]", for example: "Rating: [[5]]".

[Question]
{instruction}

[The Start of Assistant's Answer]
{response}
[The End of Assistant's Answer]
\end{lstlisting}
\end{tcolorbox}

\begin{tcolorbox}[left=0mm,right=0mm,top=0mm,bottom=0mm,boxsep=1mm,arc=0mm,boxrule=0pt, frame empty, breakable]
\textbf{Knowledge}
\small
\begin{lstlisting}
Your will be given a question, the reference answers to that question, and an answer to be judged. Your tasks is to judge whether the answer to be judged is correct, given the question and reference answers. An answer considered correct expresses or contains the same meaning as at least **one of** the reference answers. The format and the tone of the response does not matter.

You should respond in JSON format. First provide a one-sentence concise analysis for the judgement in field `analysis`, then your judgment in field `judgment`. For example,
```json
{{"analysis": <a one-sentence concise analysis for the judgement>, "judgment": <your final judgment, "correct" or "incorrect">}}
```

# Question
{instruction}

# Reference Answer
{targets}

# Answer To Be Judged
{answer_to_be_judged}
\end{lstlisting}
\end{tcolorbox}

\end{document}